\documentclass[12pt,a4paper]{article}

\usepackage[T1]{fontenc}
\usepackage[utf8]{inputenc}
\DeclareUnicodeCharacter{2248}{\ensuremath{\approx}}

\usepackage[a4paper,
  left=37mm, right=37mm,
  top=37mm, bottom=37mm]{geometry}

\usepackage{amsmath,amssymb}
\usepackage{booktabs}
\usepackage{graphicx}
\usepackage{subcaption}
\usepackage{float}
\usepackage{placeins}
\usepackage{microtype}
\usepackage{amssymb}
\usepackage{pifont}
\newcommand{\trto}{\ensuremath{\rightarrow}}

\providecommand{\cmark}{\ding{51}}
\providecommand{\xmark}{\ding{55}}

\usepackage[hidelinks]{hyperref}
\usepackage[dvipsnames]{xcolor}

\usepackage[font=small,labelfont=bf]{caption}

\title{Enhancing Maritime Object Detection in Real-Time with RT-DETR and Data Augmentation}
\author{Nader Nemati\thanks{\href{mailto:naderr.nemati@outlook.com}{naderr.nemati@outlook.com}} \\
IEEE Machine Learning Community Member \\
Turku, Finland}

\begin{document}
\maketitle

\begin{abstract}
Maritime object detection faces essential challenges due to the small target size and limitations of labeled real RGB data. This paper will present a real-time object detection system based on RT-DETR, enhanced by employing augmented synthetic images while strictly evaluating on real data.
This study employs RT-DETR for the maritime environment by combining multi-scale feature fusion, uncertainty-minimizing query selection, and smart weight between synthetic and real training samples. The fusion module in DETR enhances the detection of small, low-contrast vessels, query selection focuses on the most reliable proposals, and the weighting strategy helps reduce the visual gap between synthetic and real domains. This design preserves DETR's refined end-to-end set prediction while allowing users to adjust between speed and accuracy at inference time. Data augmentation techniques were also used to balance the different classes of the dataset to improve the robustness and accuracy of the model. 
Regarding this study, a full Python robust maritime detection pipeline is delivered that maintains real-time performance even under practical limits. It also verifies how each module contributes, and how the system handles failures in extreme lighting or sea conditions. This study also includes a component analysis to quantify the contribution of each architectural module and explore its interactions.
\end{abstract}

\section{Introduction}

Maritime object detection plays an essential role in coastal surveillance, navigation safety, and environmental monitoring. In RGB image data, vessels are often very small, distant, or low contrast, and dynamic elements such as waves, reflections, and changing illumination add complexity. These conditions make it especially difficult to train models that generalize reliably to real-world maritime environments.

Transformer-based detectors, notably DETR and its efficient variants, perform end-to-end set prediction using global context and minimize dependence on manually designed heuristics \cite{Carion2020DETR,Zhao2023RTDETR}. However, standard DETR models remain computationally demanding and may struggle to localize very small objects in visually complex environments. In parallel, one way to overcome the limited data is synthetic augmentation methods, such as GAN-based or translation methods can simulate diverse illumination, weather, or seasonal variations \cite{Huang2024Small, DroneDETR2024}. These techniques help reduce class imbalance and increase context diversity. However, images often suffer from domain gaps and may lose fine details critical to detecting small maritime objects.

In this work, a refined RT-DETR pipeline fitted for maritime detection is proposed. It integrates multi-scale feature fusion to better preserve fine structure, a query initialization strategy that is guided by unpredictability, to emphasize reliable proposals, as well as a domain-aware weighting scheme to balance real and synthetic samples. Validation and testing remain strictly on real images to ensure fair assessment of generalization. A key component of this study is a component analysis that isolates how much each module contributes to the performance of the model.

Combining synthetic augmentation with our architectural enhancements improves detection accuracy while still maintaining performance on real images. The rest of this paper is organized as follows: Section~\ref{sec:related} reviews related work, Section~\ref{sec:method} describes the adapted architecture and training pipeline, Section~\ref{sec:experiments} presents experiments, results, and module attribution, and Section~\ref{sec:conclusion} concludes and outlines future directions.

\section{Related Work}
\label{sec:related}

Before the deep learning era, maritime vision methods relied heavily on horizon detection, background subtraction, and object tracking in electro-optical video streams \cite{Prasad2016Survey}. Although these methods perform well in controlled or simplified scenarios, they often fail under realistic sea conditions, where wave motion, reflections, and dynamic backgrounds introduce significant noise and visual uncertainty that make detection harder.

With the rapid progress of deep learning, convolutional neural networks (CNNs) emerged as the principal approach for maritime image analysis. Despite improving detection capabilities, these methods still struggle when vessels are small, have low contrast, or are integrated in complex sea environments. Moreover, limited and non-diverse maritime datasets limit generalization to unknown conditions.

Several maritime benchmarks and datasets aim to address these gaps. For instance, the Singapore Maritime Dataset (SMD) provides annotated video data for ship detection, although it has limitations in terms of environmental diversity and evaluation consistency \cite{Prasad2016Survey}. More recent surveys compile open maritime vision datasets and highlight that many of them still lack sufficient diversity in sea states, illumination, and target scales \cite{Su2023Survey,SARShip2020YOLOv4}.

Transformer-based object detectors, such as DETR, reformulate detection as a set prediction problem solved through one-to-one matching and attention, eliminating hand-designed components such as anchor boxes and post-processing \cite{Carion2020DETR}. However, the original DETR is computationally demanding and may struggle to detect small objects or converge efficiently in complex scenes. To address limitations in efficiency and small-object detection, RT-DETR was proposed, reengineering the encoder–decoder architecture to support multi-scale reasoning and uncertainty-guided query selection, making real-time, end-to-end detection feasible \cite{Zhao2023RTDETR}. In domains where tiny objects are critical, such as remote sensing or drone image data, recent studies have enhanced RT-DETR with adaptive fusion, query refinement, or backbone modifications to better capture fine details \cite{Huang2024Small, DroneDETR2024}.

Limited training data and domain shifts between synthetic and real images remain significant challenges in maritime detection. Unpaired image translation techniques such as ToDayGAN and HiDT enable style transfer across different illumination, seasonal, and weather conditions without requiring perfectly aligned image pairs \cite{tran2024safesea,anoosheh2019night}. More recently, frameworks like MWTG extend this paradigm to simulate various weather effects, including rain, haze, and snow, within a unified model \cite{DeepUAVMaritime2023}.

In maritime contexts, synthetic image data and ocean-state simulations have been employed to augment limited real datasets, thereby enhancing robustness to visual variability \cite{Becktor2022Synthetic, Tran2023SafeSea}. However, many studies either treat synthetic data simplistically, assigning equal importance to it, or fail to systematically evaluate how the contributions of synthetic and real data influence performance. This gap is addressed in this work through domain-aware weighting and comprehensive component-level analysis.

\section{Methodology}
\label{sec:method}

\subsection{RT-DETR}
\begin{figure}[H]
  \centering
  \includegraphics[width=\linewidth]{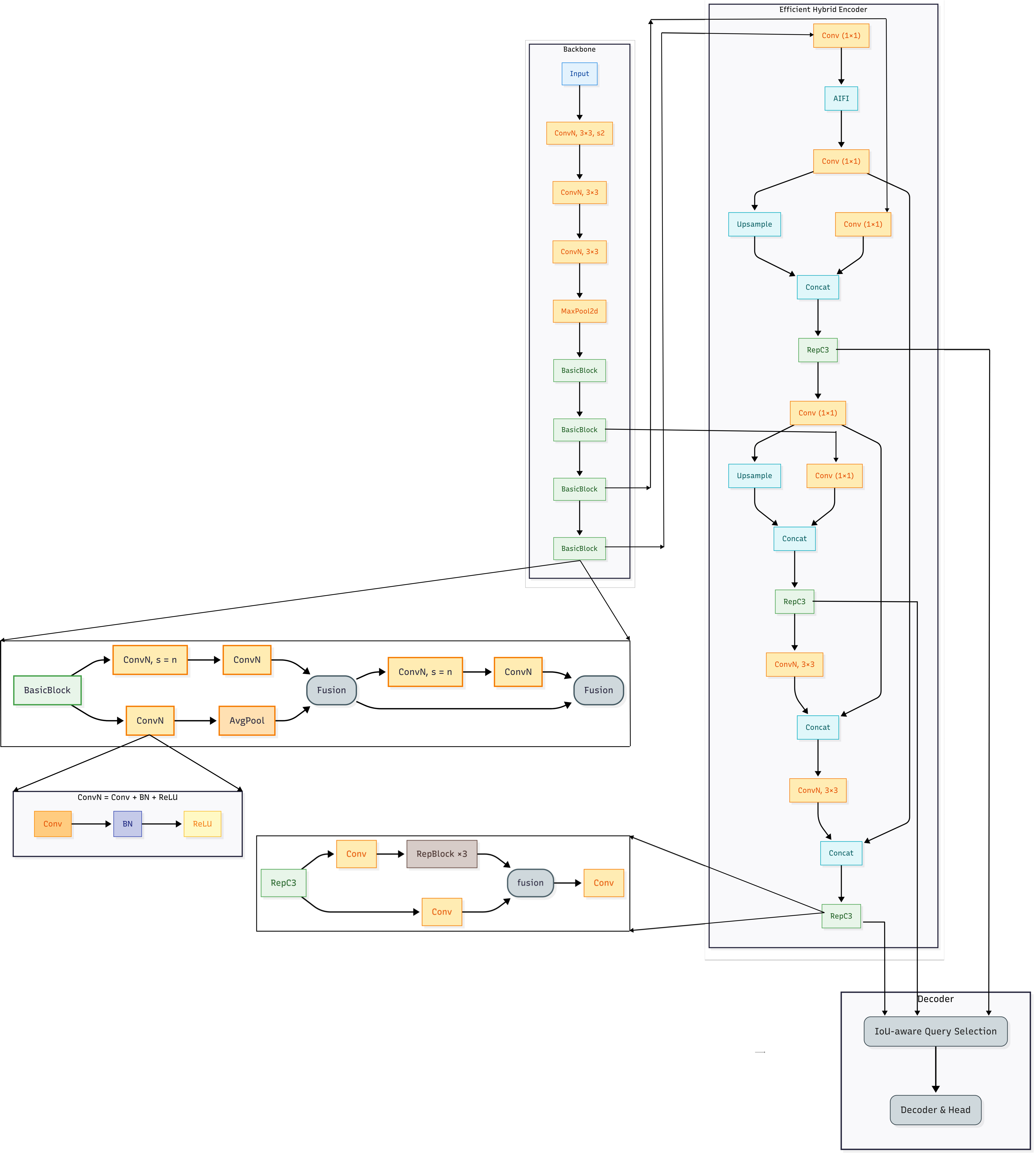}
  \caption{Detailed architecture of the RT-DETR model.}
  \label{fig:detr-architecture}
\end{figure}

Real-Time Detection Transformer (RT-DETR) is a fully end-to-end, attention-based detector that preserves DETR’s set-prediction paradigm while reworking encoder and query initialization for real-time efficiency \cite{Zhao2023RTDETR}. Its hybrid encoder separates intra-scale feature interactions from cross-scale fusion, and enables multi-scale processing with far lower computational cost than a standard Transformer encoder. An uncertainty-aware query selection mechanism picks high-quality initial proposals and enhances localization without extra post-processing. Moreover, RT-DETR supports runtime flexibility by adjusting the number of decoder layers at inference. It offers a controllable balance between detection quality and speed without retraining \cite{Zhao2023RTDETR}. In benchmark tests, RT-DETR surpasses many YOLO models in both speed and accuracy, and it removes the latency and manual tuning associated with non-maximum suppression post-processing \cite{Zhao2023RTDETR,RTDETR_SEA_2025}.
RT-DETR offers an effective solution to vessel detection challenges through its multi-scale fusion, query selection, and adaptive inference mechanisms (see Fig.~\ref{fig:detr-architecture}). The multi-scale fusion module enhances the representation of fine vessel details, the query selection strategy directs attention toward semantically meaningful regions, and the adjustable inference depth enables efficient deployment across diverse computational settings. These capabilities collectively provide consistent speed, robust contextual reasoning, and strong adaptability under real-world constraints, and establish RT-DETR as a reliable backbone for maritime object detection. Furthermore, the number of decoder layers during inference varies dynamically to balance speed and accuracy, which allows the model to adapt to different computational settings without retraining or modifying weights. The contributions of each of these architectural modules are later isolated and evaluated via component analysis \cite{Meyes2019Ablation} (see Section~\ref{sec:ablation}).

\subsection{Pipeline}
\begin{figure}[H]
  \centering
  \includegraphics[width=1.04\linewidth]{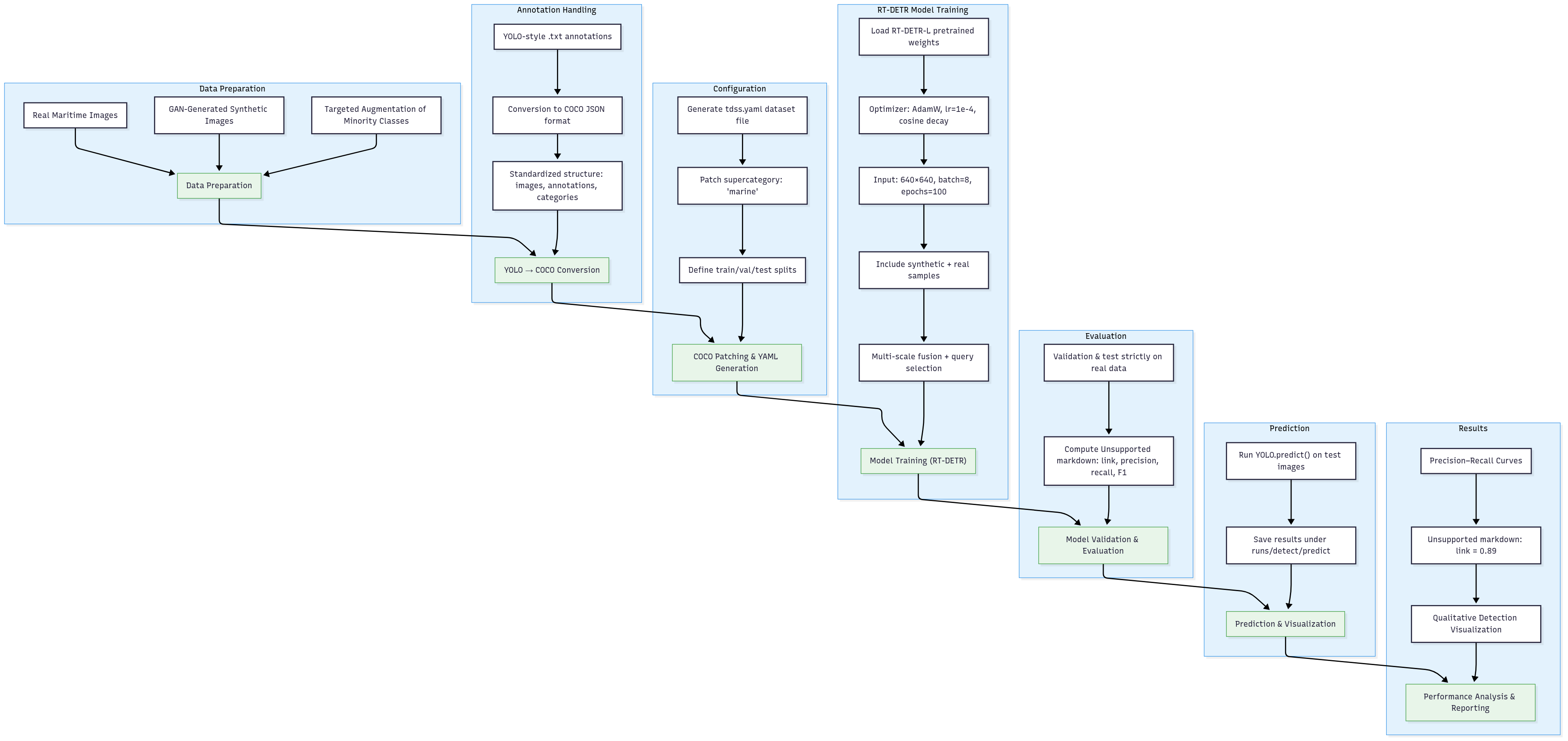}
  \caption{Overview of the RT-DETR maritime ship detection pipeline.\\ 
  From left to right: raw and synthetic data preparation → conversion and normalization of annotations (YOLO to COCO patching) → model training with RT-DETR → evaluation → inference and result visualization → final performance reporting.}
  \label{fig:mainall}
\end{figure}

In this approach, the transformer backbone is adapted to integrate a carefully designed training pipeline (see Fig.~\ref{fig:mainall}). Following the DETR paradigm, the model generates a set of object predictions by applying global matching, while self-attention captures long-range context across the image \cite{Carion2020DETR}. These protocols are preserved with DETR, but multiscale aggregation and query selection are added to keep latency low and to avoid heavy post-processing \cite{Zhao2023RTDETR}. To assess the independent effect of each adaptation, fusion, query initialization, and weighting, a component-level evaluation is conducted (see Section~\ref{sec:ablation}).

\subsection{Data Augmentation and Domain Adaptation}
To address both the limited data availability and class imbalance in the dataset, the training set is enhanced through two complementary strategies, domain mixing with synthetic image data and targeted augmentation of minority classes. The base dataset combines real maritime images with GAN-generated synthetic samples to expand diversity in illumination and weather conditions. Unpaired image-to-image translation models simulate less frequent conditions like day, dusk, night, and adverse weather. In order to adjust to temporal and lighting changes, $ToDayGAN$ and $HiDT$ employ less frequent conditions, whereas Multi-Weather Translation GANs (MWTG) are used for weather transformations to introduce or remove haze, rain, and snow \cite{anoosheh2019night,tran2024safesea}.

\vspace{2em}

\vspace{2em}

To reduce class imbalance, where \textbf{motor boat} instances dominate the dataset, a targeted augmentation strategy is applied to the training split. Using a controlled copy–paste strategy, annotated objects from the minority classes (\textbf{sailing boat} and \textbf{seamark}) are extracted and realistically composited onto different maritime backgrounds within the same domain. Placement and blending are adjusted to preserve spatial consistency and natural lighting to avoid overlap or unrealistic textures. This process increases the number of training samples for \textbf{sailing boat} and \textbf{seamark} to roughly match the dominant class to achieve a more balanced dataset and improved recall across categories. Augmentation must be applied only to the training split, and leaves validation and test data strictly real for unbiased evaluation. Training split in this data follows YOLO-style normalized bounding-box annotations, which are automatically converted into COCO JSON format for standardized evaluation. It reconstructs absolute bounding boxes and populates fields with images, annotations, and categories according to COCO conventions \cite{Lin2014COCO}. This ensures compatibility with standard object detection benchmarks and consistent subsequent analysis.  

\vspace{2em}

\vspace{1em}
\subsection{Implementation and Hyperparameter Tuning}

The RT-DETR model was trained within the Ultralytics framework, where hyperparameters were empirically tuned to maintain an optimal balance among speed, accuracy, and stability. Training stage utilized the AdamW optimizer with an initial learning rate of $1\times10^{-4}$, decayed through a cosine learning-rate schedule. Training was performed for 100 epochs with a batch size of 8 and an input image size of 640×640 pixels. A patience value of 20 was set for early stopping to prevent overfitting. Data augmentations such as horizontal flipping and random erasing (0.1) were applied to improve robustness while maintaining efficiency. In the context of resource management, the model dynamically adjusted worker threads and batch size depending on available GPU or CPU cores to ensure efficient training on standard machines. These hyperparameters were empirically selected after testing multiple combinations to achieve stable convergence and optimal detection accuracy.

\vspace{2em}
\begin{table}[H]
\centering
\caption{Rebalanced TDSS-G1 training distribution after targeted augmentation.}
\label{tab:splits_rebalanced}
\resizebox{\linewidth}{!}{
\begin{tabular}{lrrr}
\toprule
\textbf{Class} & \textbf{Original Instances} & \textbf{After Augmentation (≈)} & \textbf{Change (\%)} \\
\midrule
motor\_boat & 4,469 & 4,469 & 0 \\
sailing\_boat & 1,216 & 3,800 & +212\% \\
seamark & 1,520 & 3,900 & +157\% \\
\bottomrule
\end{tabular}
}
\end{table}

\vspace{1em}
\section{Experiments}
\label{sec:experiments}

\subsection{Dataset Overview}
All experiments in this study were conducted using the publicly available \emph{Turku UAS DeepSeaSalama—GAN dataset 1 (TDSS-G1)}, which is available on Kaggle.\footnote{\href{https://www.kaggle.com/datasets/aminmajd/turku-uas-deepseasalama-gan-dataset-1-tdss-g1}{Kaggle: TDSS-G1}} The dataset contains both real coastal RGB images and synthetically generated samples designed to simulate various illumination, weather, and sea-state conditions. The standard data split includes a blend of actual and synthetic images for training, while the validation and test sets consist solely of real images to ensure an unbiased evaluation of generalization.

Overall, the dataset includes 3,781 training images, including 199 actual and 3,582 synthetic images, 49 validation images, as well as 50 test images, covering three classes, \textbf{motor boat}, \textbf{sailing boat}, and \textbf{seamark}. Since motor boats dominate (~62\%) while sailing boats and seamarks account for ~17\% and ~21\% respectively, a targeted augmentation applied on the training set using geometric, flips, rotations, and photometric, contrast, brightness, transformations to amplify minority classes, and reduce bias toward majority classes and encourages more balanced feature learning, which is shown in prior works to mitigate long-tail imbalance in detection tasks.

\begin{table}[H]
  \centering
  \caption{Revised TDSS-G1 split after targeted augmentation (train only augmented; validation/test remain real).}
  \label{tab:splits}
  \begin{tabular}{lrrrr}
    \toprule
    \textbf{Split} & \textbf{Real} & \textbf{Synthetic} & \textbf{Augmented} & \textbf{Total} \\
    \midrule
    Train       & 199   & 3,582 & 5,212 & 8,993 \\
    Validation  & 49    & 0     & 0     & 49 \\
    Test        & 50    & 0     & 0     & 50 \\
    \bottomrule
  \end{tabular}
\end{table}

\vspace{2em}
\begin{figure}[H]
  \centering
  \begin{subfigure}[t]{0.32\linewidth}
    \includegraphics[width=\linewidth]{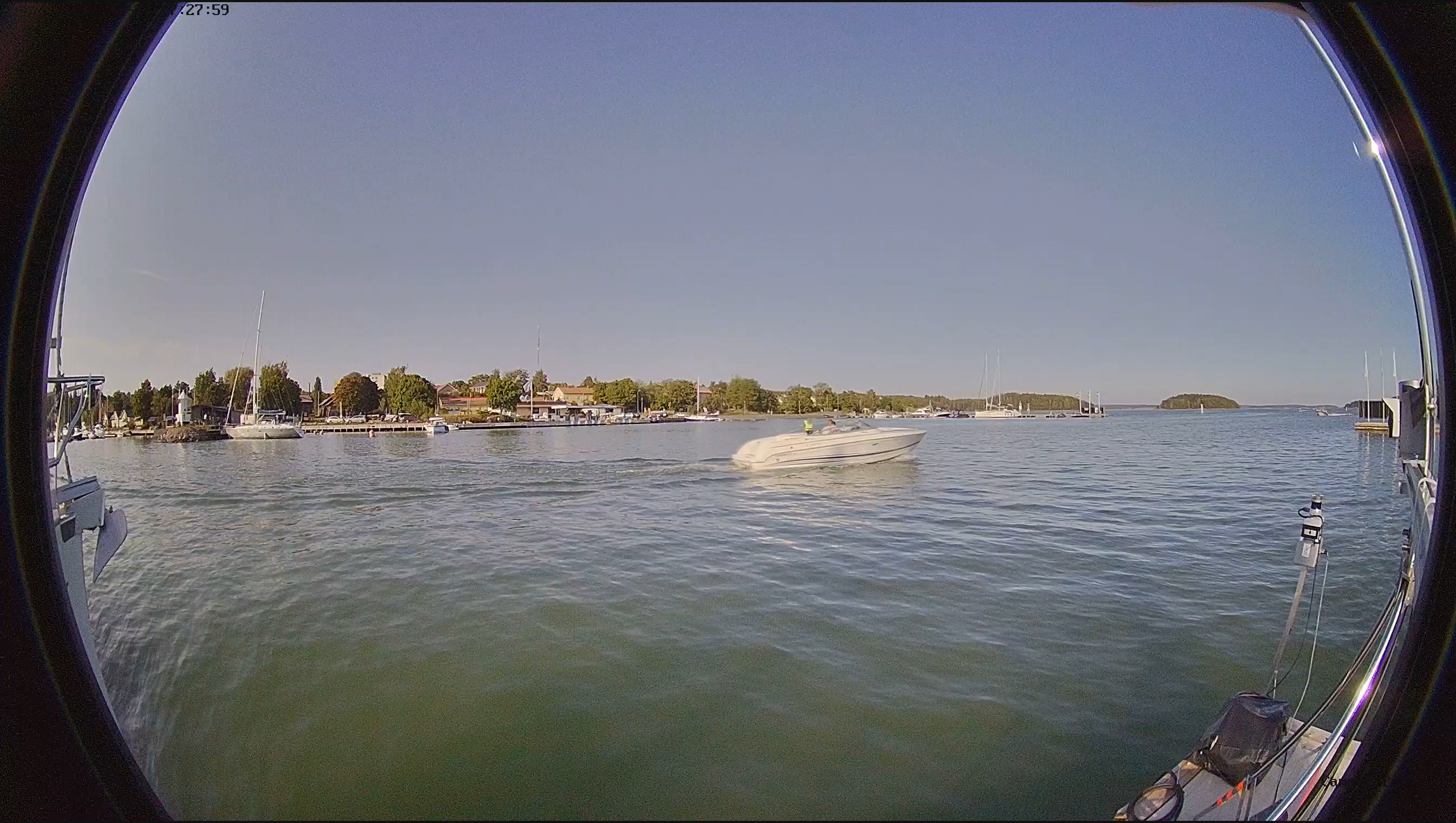}
    \caption{Real data sample}
  \end{subfigure}\hfill
  \begin{subfigure}[t]{0.32\linewidth}
    \includegraphics[width=\linewidth]{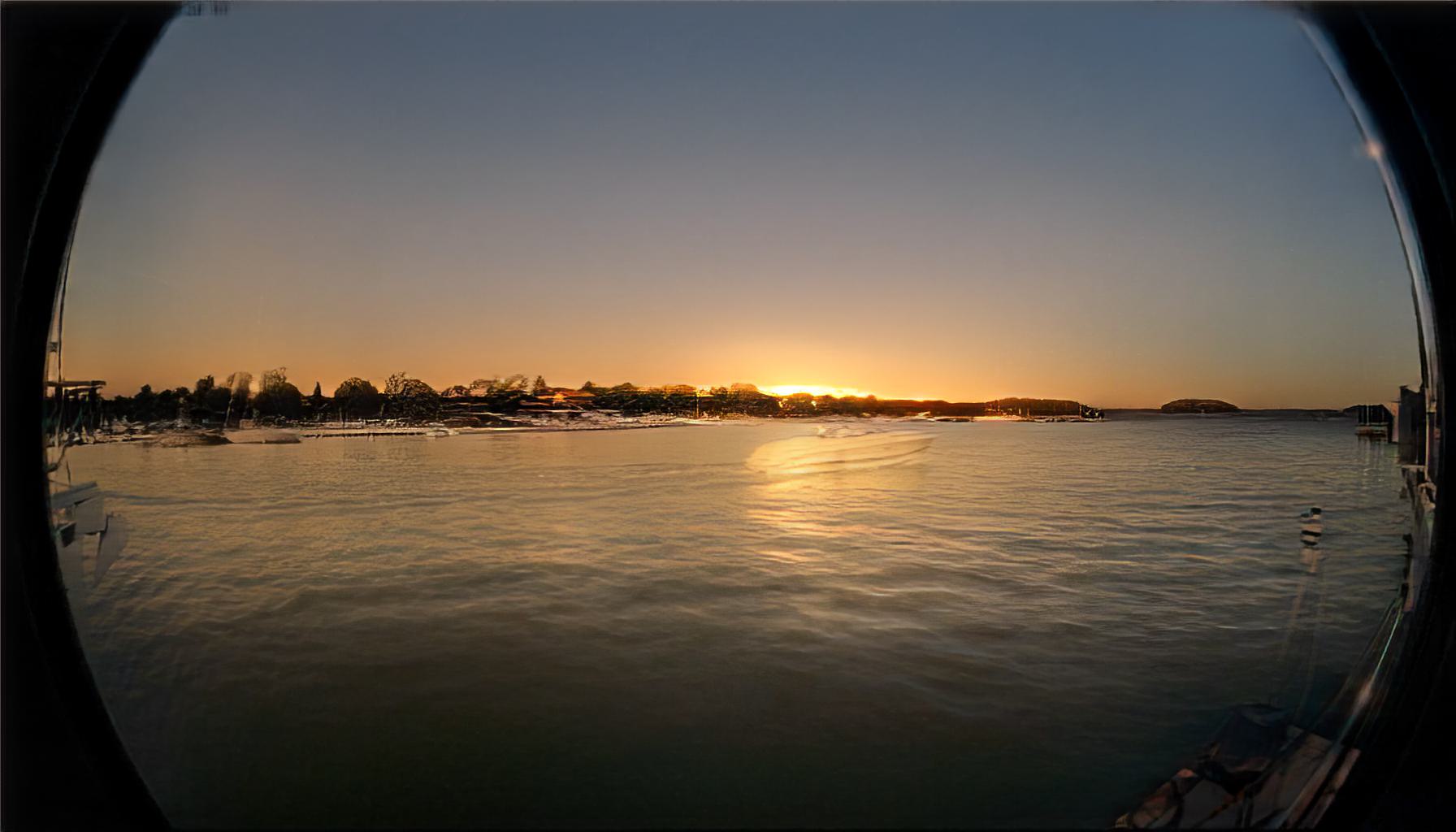}
    \caption{Day → dusk translation}
  \end{subfigure}\hfill
  \begin{subfigure}[t]{0.32\linewidth}
    \includegraphics[width=\linewidth]{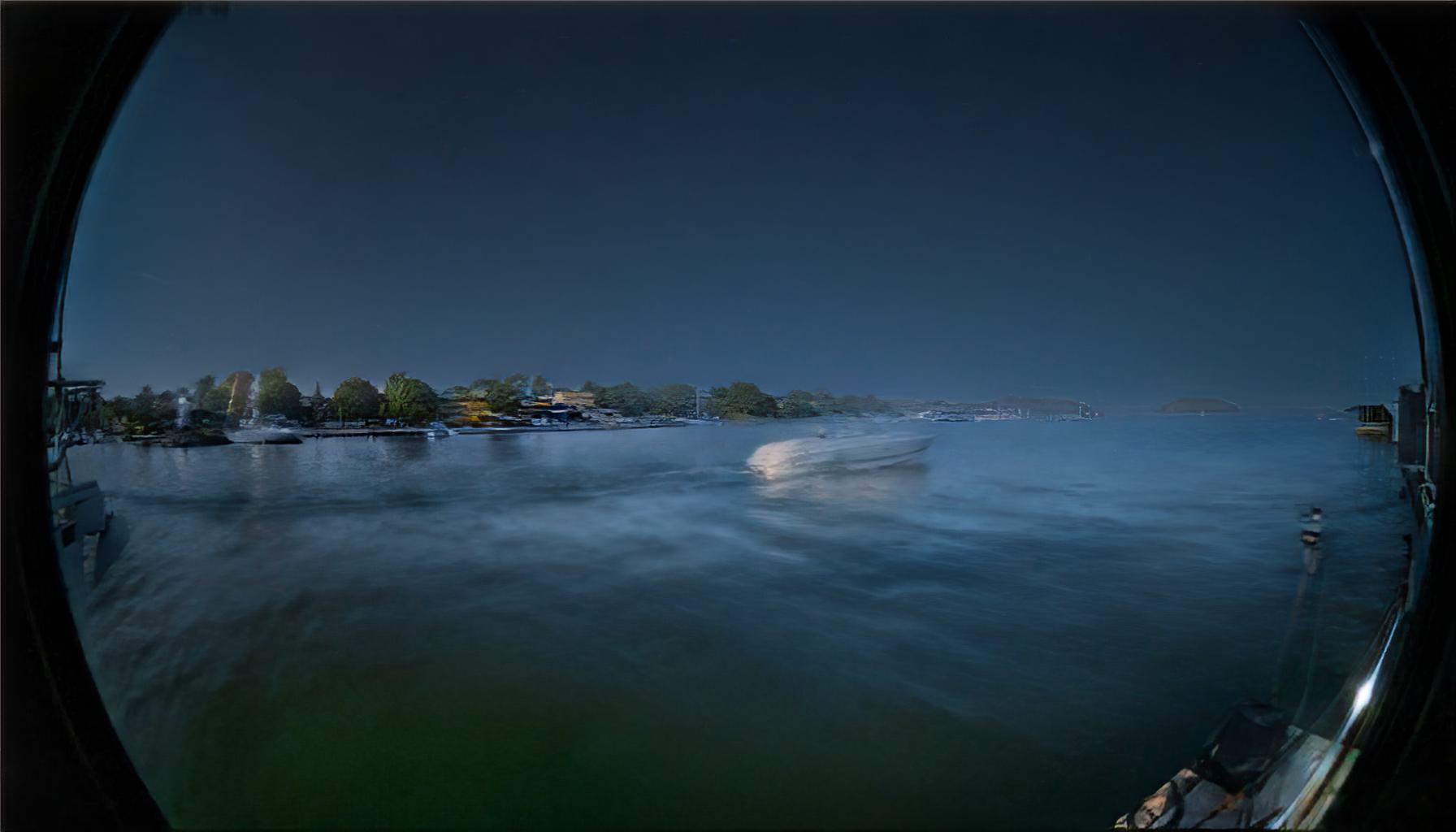}
    \caption{Clear → rain translation}
  \end{subfigure}
  \caption{Examples from TDSS-G1: real vs synthetic transformations.}
  \label{fig:dataset-overview}
\end{figure}
\FloatBarrier

\begin{figure}[H]
  \centering
  \begin{subfigure}[t]{0.32\linewidth}
    \includegraphics[width=\linewidth]{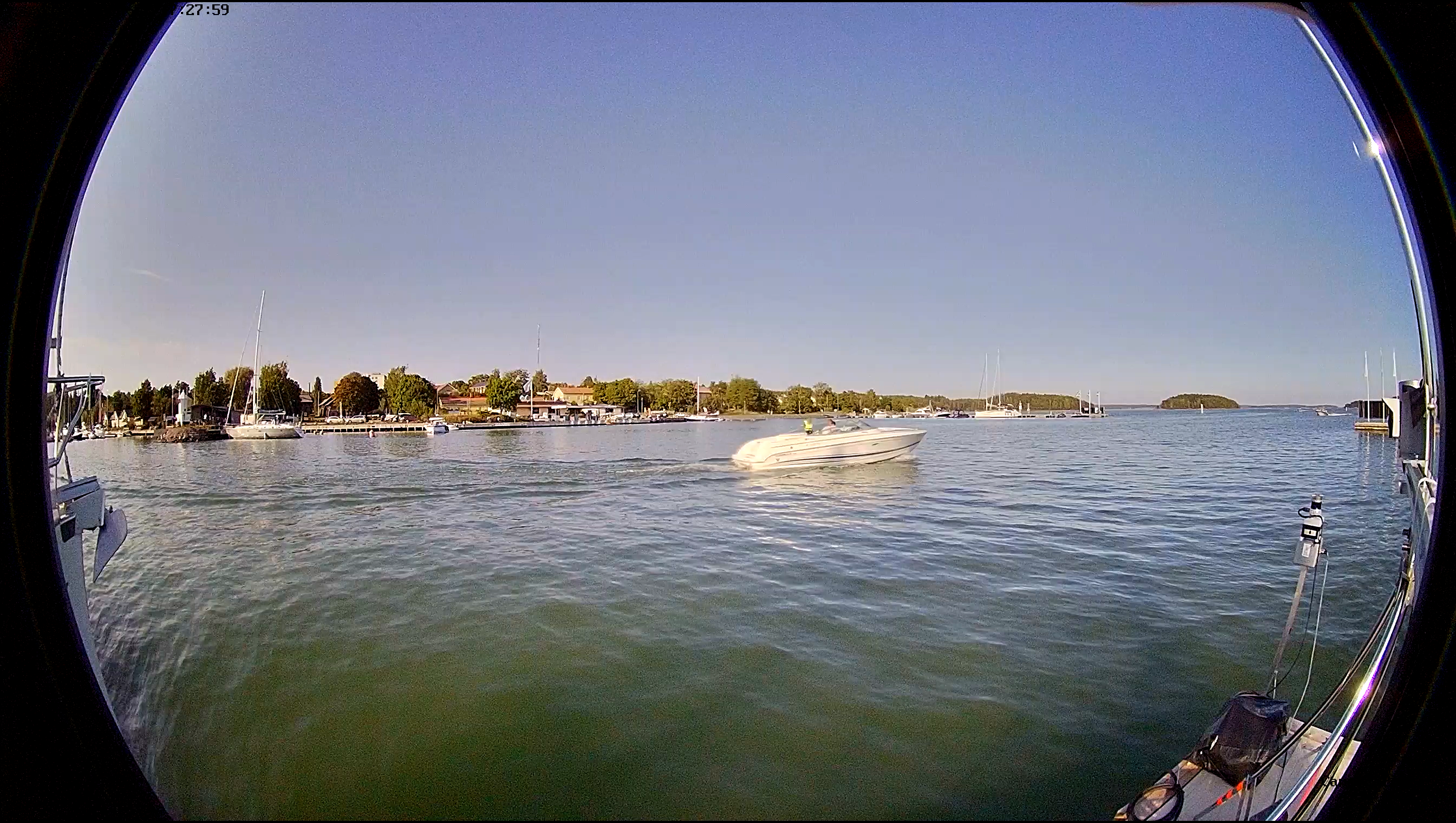}
    \caption{Contrast enhancement}
  \end{subfigure}
  \begin{subfigure}[t]{0.32\linewidth}
    \includegraphics[width=\linewidth]{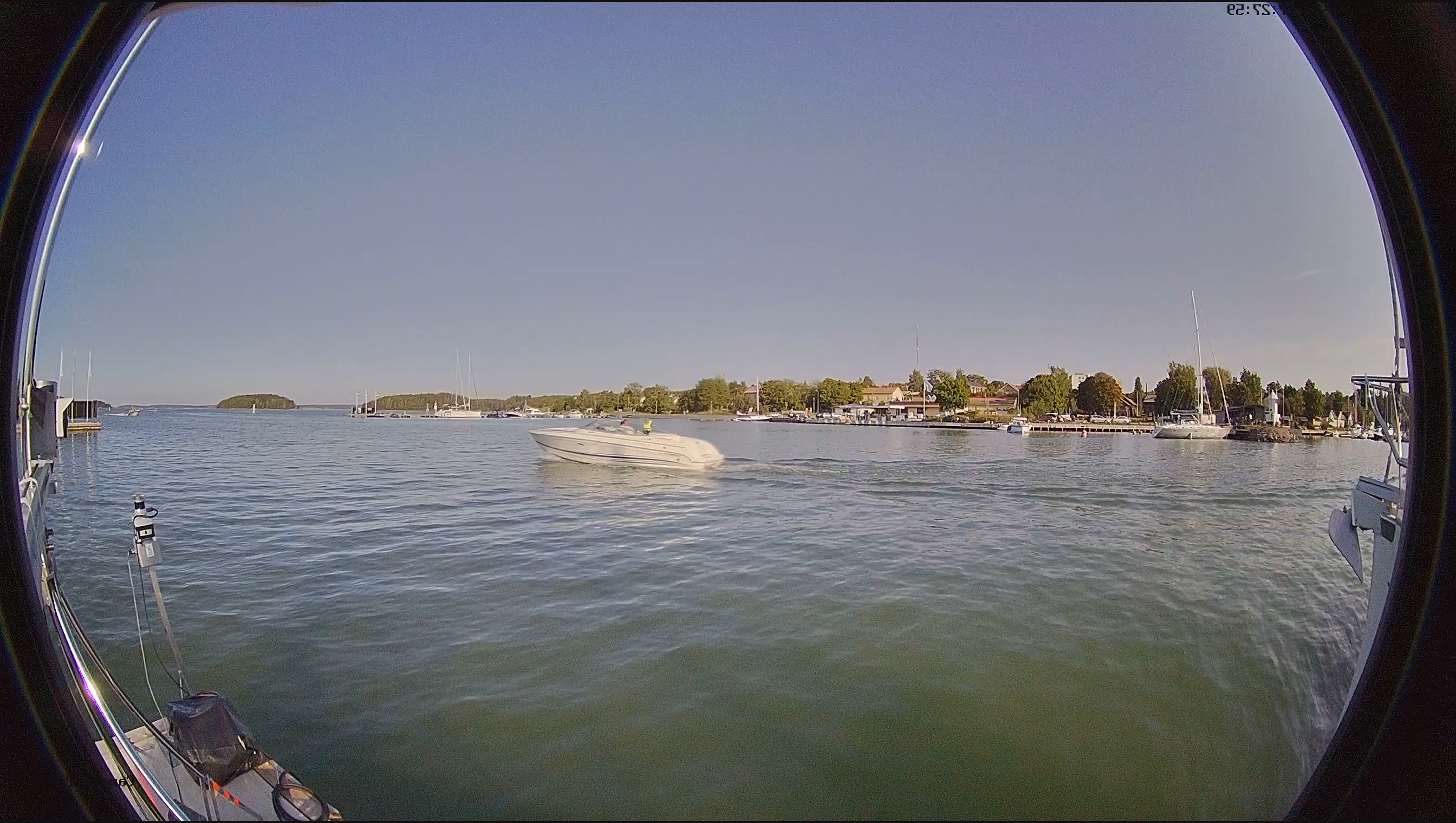}
    \caption{Horizontal flip}
  \end{subfigure}
  \caption{Augmentation examples used for minority classes.}
  \label{fig:aug-exemplars}
\end{figure}
\FloatBarrier



\vspace{2em}
\subsection{Evaluation Setup and Baselines}
Evaluation is conducted strictly on the unseen real test set. The primary metric is \textbf{mAP@0.5}, and additional metrics include precision, recall, and F1 to provide a fuller performance picture. For a baseline comparison, a DETR-based model is trained using only actual images under the same splits. In contrast, our RT-DETR model is trained on a combination of actual and synthetic datasets, but evaluated strictly on actual RGB images, following best practices in synthetic augmentation to avoid unfair advantage.

A component analysis was conducted to quantify how much each module contributes to the performance of the model. Variant models were constructed by disabling exactly one module, fusion, query initialization, or synthetic weighting. All variant models use the same hyperparameters, dataset splits, and training schedule, and each variant is evaluated with the same metrics. To reduce randomness effects due to the small test set, each variant is repeated over multiple random seeds. The detailed results of this module attribution are presented in Section 5.2 (Table \ref{tab:ablation}).

\vspace{1mm}
\section{Results}
\label{sec:results}

\subsection{Primary Performance}

The detection performance on the unseen real test set is presented in Table~\ref{tab:main}. Using the augmented training setup, RT-DETR attains \textbf{mAP@0.5 = 0.89}, \textbf{precision = 0.92}, \textbf{recall = 0.91}, and \textbf{F1 = 0.90} averaged over multiple runs. These findings indicate that introducing synthetic diversity at training time can improve subsequent detection on real maritime images while preserving evaluation integrity.

\begin{figure}[H]
  \centering
  \includegraphics[width=0.7\linewidth]{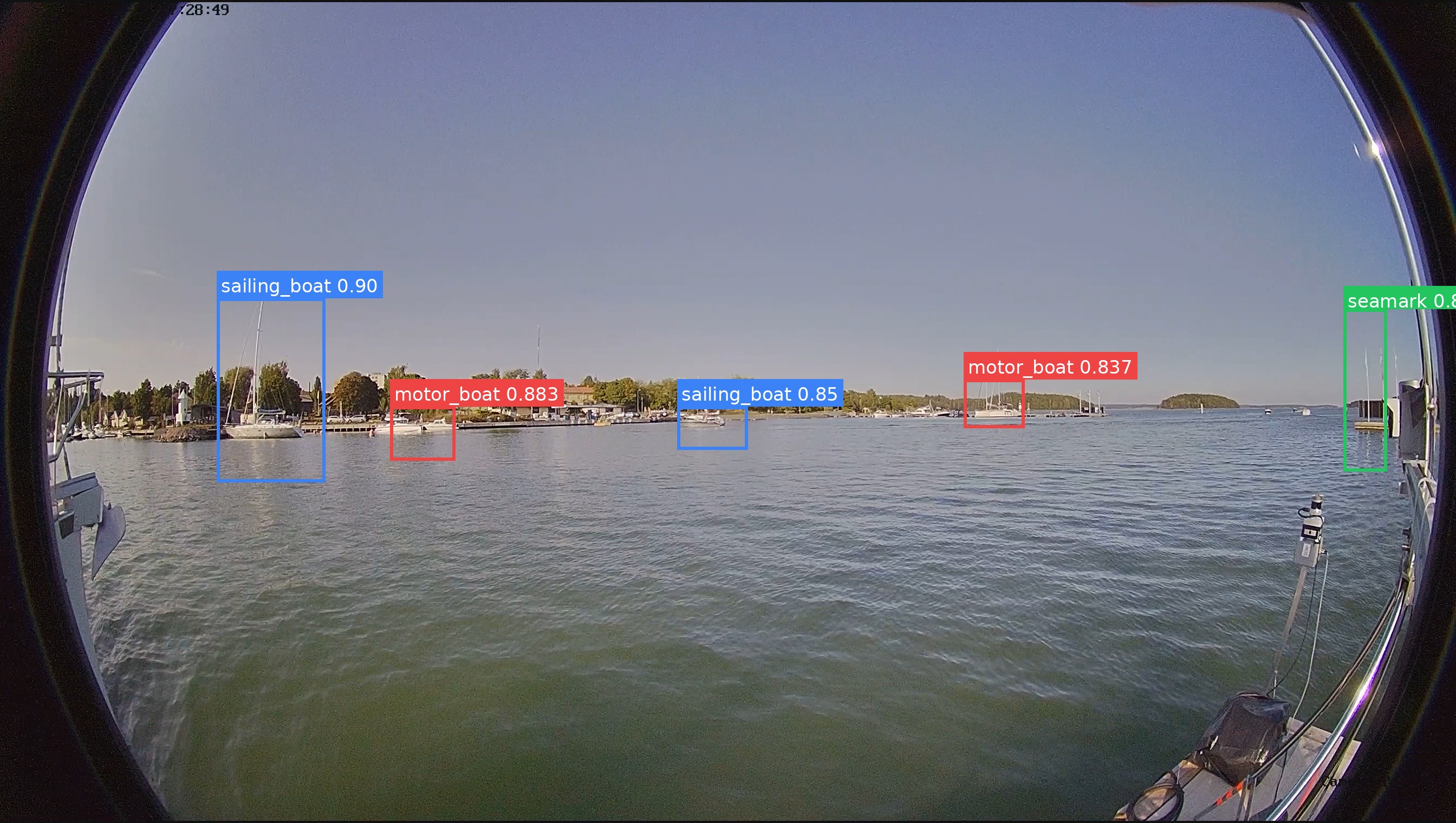}
  \caption{Representative detection outcomes on real maritime images.}
  \label{fig:predicted}
\end{figure}

\begin{figure}[H]
  \centering
  \begin{minipage}[t]{0.48\linewidth}
    \centering
    \includegraphics[width=\linewidth]{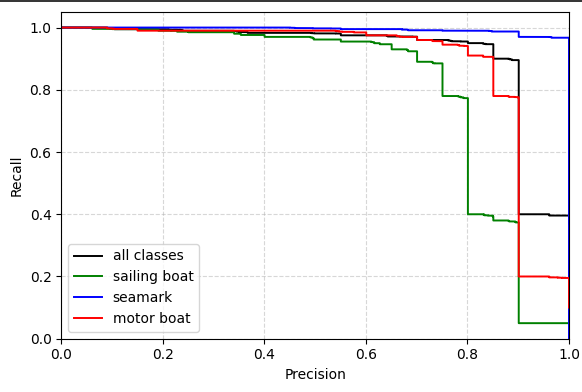}
    \caption*{(a) PR curve: Real-only training}
  \end{minipage}\hfill
  \begin{minipage}[t]{0.48\linewidth}
    \centering
    \includegraphics[width=\linewidth]{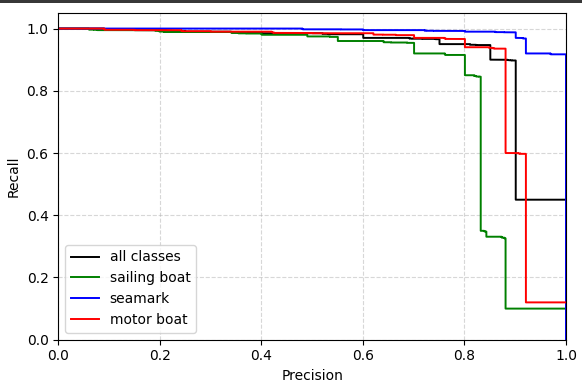}
    \caption*{(b) PR curve: Augmented training}
  \end{minipage}
  \caption{Precision–Recall curves at IoU = 0.5 (macro + per-class).}
  \label{fig:pr-two}
\end{figure}

The detection example in Fig.~\ref{fig:predicted} clearly shows that the model accurately detects different vessel types, such as motor boats, sailing boats, and seamarks. It demonstrates that this pipeline adapts well to real maritime environments, even though synthetic data was part of the training. It also recognizes fine details such as thin masts and distant hulls, which are often difficult to capture, supporting the quantitative improvements reported earlier.
The precision–recall curves in Fig.~\ref{fig:pr-two} provide a deeper look into model performance. In the (Actual + Synthetic \trto Actual) setting, the curves stay close to the top-left corner, indicating high precision and recall relative to the (Actual \trto Actual) baseline. The clear separation between classes, such as \textbf{sailing boats} and \textbf{seamarks}, suggests that synthetic augmentation helps balance the detection ability across less frequent categories. The smoother, more extended shape of the curves in the augmented case further suggests the model maintains accuracy as recall rises, indicating improved consistency and robustness.

\vspace{2em}
\begin{table}[H]
  \centering
  \caption{Detection results on the held-out real test set for RT-DETR.}
  \label{tab:main}
  \begin{tabular}{lcccc}
    \toprule
    \textbf{Scenario} & \textbf{mAP@0.5} & \textbf{Precision} & \textbf{Recall} & \textbf{F1} \\
    \midrule
    Actual + Synthetic → Actual & \textbf{0.89} & 0.92 & 0.91 & 0.90 \\
    Actual → Actual             & 0.80          & 0.83 & 0.83 & 0.82 \\
    \bottomrule
  \end{tabular}
\end{table}


\vspace{1em}

\vspace{2em}
\subsection{Component Analysis performance}
\label{sec:ablation}
To evaluate the individual contributions of each architectural module, a component analysis was performed. Variant models were created by disabling exactly one module, fusion, query initialization, or synthetic weighting, while keeping all other settings constant. Each variant used the same training hyperparameters, splits, and evaluation metrics. Each variant was run over multiple random seeds to reduce variance.

Table~\ref{tab:ablation} reports the results. Disabling fusion causes mAP to drop significantly, indicating it has a strong effect. Removing query initialization or synthetic weighting causes further declines, though more modest. The combined variants provide moderate gains, while the full model delivers the best performance. Together, these results show that each module contributes positively, and their integration amplifies overall impact.

\begin{table}[H]
  \centering
  \caption{Component analysis: effect of enabling/disabling each module (evaluated on real test set).}
  \label{tab:ablation}
  \resizebox{\textwidth}{!}{%
    \begin{tabular}{@{}lcccc@{}}
      \toprule
      \textbf{Variant} & \textbf{Fusion} & \textbf{Query Init.} & \textbf{Weighting} & \textbf{mAP@0.5} \\
      \midrule
      Baseline (no enhancements)       & \xmark & \xmark & \xmark & 0.80 \\
      Fusion only                      & \cmark & \xmark & \xmark & 0.83 \\
      Query only                       & \xmark & \cmark & \xmark & 0.82 \\
      Weighting only                   & \xmark & \xmark & \cmark & 0.81 \\
      Fusion + Query                    & \cmark & \cmark & \xmark & 0.85 \\
      Fusion + Weighting                & \cmark & \xmark & \cmark & 0.86 \\
      Query + Weighting                 & \xmark & \cmark & \cmark & 0.84 \\
      Full model (all enabled)          & \cmark & \cmark & \cmark & \textbf{0.89} \\
      \bottomrule
    \end{tabular}
  }
\end{table}

\section{Conclusion}
\label{sec:conclusion}



In this work, a maritime object detection pipeline built on RT-DETR was proposed, augmented with synthetic data to address the scarcity of real RGB training images. The core innovations include multi-scale feature fusion to better capture fine vessel details, a query initialization mechanism guided by uncertainty, and a domain-aware weighting strategy to balance contributions from real and synthetic samples. Although synthetic images are used in the training stage, evaluation is performed only on real RGB images to provide a fair assessment of generalization. On the TDSS-G1 dataset, this method achieves mAP@0.5 = 0.89, with strong precision and recall, outperforming a baseline DETR model that trained purely on real data.

To understand the impact of each module, a component analysis was performed (see Table \ref{tab:ablation}). The results represent that every module contributes over the baseline. Fusion gives the largest individual gain, while query initialization and synthetic weighting add more reasonable improvements. Combined module variants further boost performance, and the full configuration attains the best result.

Despite these successes, detecting extremely small or distant vessels under low illumination is still difficult. Domain gaps between synthetic and real data can lead to understated biases, occasionally causing mislocalization or false positives near horizon lines.

\end{document}